\documentclass[journal]{IEEEtran} 
\usepackage{newtxtext,newtxmath}
\usepackage[T1]{fontenc}
\usepackage{ae,aecompl}
\usepackage{multicol} 
\usepackage{graphicx}
\usepackage{setspace}
\usepackage{tocloft}
\usepackage[utf8]{inputenc}
\usepackage{datetime}
\usepackage{array}
\usepackage{mathtools}
\usepackage{caption}
\usepackage{subcaption}
\usepackage{graphicx}
\usepackage{float}
\usepackage{wrapfig}
\usepackage{floatflt}
\usepackage{xcolor}     % Colored text 
\usepackage{comment}    % Commenting text using begin and end commands 
\usepackage{breqn}      % Break equation
\usepackage{lipsum}     % Figures across two columns

\begin{document}
% Title of the paper, and the short title which is used in the headers.
% Keep the title short and informative.

\title{Uniqueness of Iris Pattern Based on AR Model} 
%\title{Bad title but contains all important words: An AR model based Sphere packing and Daugman-like bounds for the analysis of the maximum population based on Iris Pattern} 
% The list of authors, and the shortlist which is used in the headers.
% If you need two or more lines of authors, add an extra line using \newauthor
\author{Katelyn M. Hampel, Jinyu Zuo, Priyanka Das, \\ Natalia A. Schmid, Stephanie Schuckers, Joseph Skufca, and Matthew C. Valenti 
\\
% List of institutions
%Lane Department of Computer Science and Electrical Engineering, \\ West Virginia University \\ 
% Morgantown, WV 26506, USA \\
\today
}

% These dates will be filled out by the publisher
%\date{Accepted XXX. Received YYY; in original form ZZZ}

% Enter the current year, for the copyright statements etc.
%\pubyear{2016}

% Don't change these lines

%\begin{multicols}{2}
%\label{firstpage}
%\pagerange{\pageref{firstpage}--\pageref{lastpage}}
\renewcommand{\cftdotsep}{\cftnodots}
\cftpagenumbersoff{figure}
\cftpagenumbersoff{table}

\maketitle

\begin{abstract}
% Original
% The evaluation of iris uniqueness is important in analyzing the capabilities and limitations of an iris recognition system. Daugman’s approach to iris uniqueness is perhaps the most accepted methodology proposed thus far. Daugman describes uniqueness as the ability of an iris recognition system to enroll more and more classes with the probability of collision between new and enrolled classes near zero. His approach creates unique IrisCode templates for each iris class contained in the system and evaluates the sustainable population with a fixed Hamming distance between each iris class. In our previous work~\cite{Hampel2021}, we use Rate-Distortion Theory (limits of error-correction codes) to establish bounds on the maximum possible population of iris classes that Daugman’s IrisCode can support under the constraint of a fixed Hamming Distance between codewords. As an extension of that work, we propose a new methodology to evaluate the scalability of an iris recognition system, while also measuring iris quality, using a Sphere Packing bound for Gaussian codewords and a similar approach to Daugman that uses relative entropy as a distance measure between iris classes. The methodology is illustrated on two small datasets of iris images, and the sustainable maximum population is demonstrated for each based upon the image quality. From these illustrations, we hope to help researchers understand the limitations present in their recognition system depending on the quality of their iris database. 

% Revised (by mcv on 6/12/23)
The assessment of iris uniqueness plays a crucial role in analyzing the capabilities and limitations of iris recognition systems. Among the various methodologies proposed, Daugman's approach to iris uniqueness stands out as one of the most widely accepted. According to Daugman, uniqueness refers to the iris recognition system's ability to enroll an increasing number of classes while maintaining a near-zero probability of collision between new and enrolled classes. Daugman's approach involves creating distinct IrisCode templates for each iris class within the system and evaluating the sustainable population under a fixed Hamming distance between codewords. In our previous work~\cite{Hampel2021}, we utilized Rate-Distortion Theory (as it pertains to the limits of error-correction codes) to establish boundaries for the maximum possible population of iris classes supported by Daugman's IrisCode, given the constraint of a fixed Hamming distance between codewords. Building upon that research, we propose a novel methodology to evaluate the scalability of an iris recognition system, while also measuring iris quality. We achieve this by employing a sphere-packing bound for Gaussian codewords and adopting a approach similar to Daugman's, which utilizes relative entropy as a distance measure between iris classes. To demonstrate the efficacy of our methodology, we illustrate its application on two small datasets of iris images. We determine the sustainable maximum population for each dataset based on the quality of the images. By providing these illustrations, we aim to assist researchers in comprehending the limitations inherent in their recognition systems, depending on the quality of their iris databases.

\end{abstract}

% Include a list of up to six keywords after the abstract

% Include email contact information for corresponding author
%{\noindent \footnotesize\textbf{*}Katelyn M. Hampel,  \linkable{kmhampel@mix.wvu.edu} }

%\begin{spacing}{1}   % use double spacing for rest of manuscript

\begin{spacing}{1}   % use double spacing for rest of manuscript
%%%%%%%%%%%%%%%%%%%%%%
\section{Introduction}
\label{introduction}
%%%%%%%%%%%%%%%%%%%%%%
The uniqueness of iris biometrics and methods to evaluate it have been central themes of multiple publications~\cite{Adler2006,Bolle2004,Schmid2008,Yoon2005}.  Daugman's approach to iris uniqueness is perhaps the most accepted methodology proposed thus far. Daugman's earlier publications~\cite{Daugman2003, Daugman2004, Daugman2006} define uniqueness as an ability of an iris recognition system to enroll more and more classes with the probability of collision between new and enrolled classes near zero. Over the years, his view on iris uniqueness has changed. His latest definition of iris uniqueness (see \cite{Daugman2016, Daugman2021} for details) is based on the analysis of a closed iris biometric system. Given an iris database of $M$ enrolled classes, uniqueness is quantified by evaluating the chance that iris images from any two randomly selected classes match. Both definitions of uniqueness were applied to assess the uniqueness of IrisCode~\cite{Daugman2003,Daugman2021,IREX-III}.         

Daugman's analysis is applicable to IrisCodes where iris templates are composed of bits. Then the distance between iris templates is conveniently measured by calculating Hamming distance. It is shown by Daugman that a binomial probability density function (pdf) with approximately 249 degrees of freedom can be fitted into the histogram of imposter Hamming distances which leads to a new interpretation of the problem of finding the maximum population size of IrisCode. One can think of the existence of a mapping from the space of IrisCode to a new space of binary codewords. In the new space, each codeword comprises 249 independent bits, and the Hamming distance between any two codewords is equal to 249.  Given that each class is represented by a unique binary codeword with the properties above, an elegant performance analysis invoking limits of error correction codes and an asymptotic case of pairwise binary detection problems easily follows.   
%describing it   and bits  \color{purple}This allows us to restate the problem of finding the maximum population size of IrisCode as a problem with iris temples replaced by a set of binary codewords of length 249 composed of independent bits, with each pair of codewords being about 249 Hamming units apart. (Reword) \color{black}

In this work, we turn to another interpretable model for iris data, a Gaussian model. To prepare the ground for the analysis of iris uniqueness, we appeal to Auto-Regressive (AR) model for the data~\cite{Hampel2022}. When driven by a white Gaussian noise  process, the AR model generates a stationary Gaussian random process unique for each iris class. Given a random description of each iris class, the problem of iris recognition is restated as an M-ary detection problem which is further simplified by replacing it with a union of pairwise binary detection problems. The log-likelihood ratio test statistic in asymptotic form is implemented for each pair of classes, which, when averaged over a large number of images per class, can be substituted with an estimate of the relative entropy between a pair of jointly Gaussian probability density functions each with zero mean and estimated power spectral density. It is further justified that the histogram of the pairwise log-likelihood ratios can be fitted with a chi-square curve. Its degrees of freedom and the scaling factor are determined by minimizing the least square distance (it performs similarly to the chi-square test) over a broad range of the two parameters. Similar to how Daugman interprets 249 degrees of freedom of the fitted Binomial curve as the length of virtual binary codewords with independent bits, we interpret the degrees of freedom of the fitted chi-square curve as the length of virtual Gaussian codewords and the scaling parameter of the fitted chi-square curve as the variance of each codeword entry. The entries of the codewords are independent and identically distributed. 

Given the fitted chi-square model, two different approaches to quantify the uniqueness of iris biometrics are presented: 
%Bound on Probability of error. How many classes can be recognized with probability of error below a certain value?   
(1) A sphere packing argument for the Gaussian source is applied first. The log-likelihood ratio statistic and the estimate of the relative entropy between two iris classes are N-Erlang distributed. This distribution can be alternatively thought of as being due to a sum of $2N$ squared independent identically distributed real-valued Gaussian random variables. We can fit a chi-square distribution with $2N$ degrees of freedom, then apply a sphere packing argument to find the dependence of the maximum population and the distortion in the data.  
(2) It is followed by a Daugman-like bound where False Match Rate (FMR) \cite{Daugman2004} is replaced by estimated relative entropy or by the log-likelihood statistic averaged over multiple images of each of the two classes. 

The rest of the paper is organized as follows. Section \ref{sec:model} presents assumptions, model, and theory. Section \ref{sec:implementation} explains how we can arrive at the theorized model in practice and presents the illustrative results of performance analysis including a sphere packing bound and a Daugman-like bound in application to two iris datasets.  Section \ref{sec:summary} summarizes the main points of the work. 

%%%%%%%%%%%%%%%%%%%%%%%%%%%%%%%%%%%%%
\section{Theory, Model, and Analysis}
\label{sec:model}
%%%%%%%%%%%%%%%%%%%%%%%%%%%%%%%%%%%%%
%
%
Assume that each iris class can be described by a piece-wise stationary Gaussian random process and that enrollment data of each iris class is a finite sample realization of a class random process. Auto-regressive (AR) process is an example of a stationary random process that can be used to model iris data.  With a Gaussian random process for each iris class, we can state the problem of iris recognition as an M-ary detection problem and then apply a variety of analytical tools to analyze the performance of a large iris biometric population.     

%%%%%%%%%%%%%%%%%%%%%%%%%%%%%%%%%%%%%%%%%%%%%%%%
\subsection{AR model for vectorized iris images} 
\label{sec:ARMA_model}
%%%%%%%%%%%%%%%%%%%%%%%%%%%%%%%%%%%%%%%%%%%%%%%%
%
Let $M$ be the number of enrolled iris classes and $N$ be the number of images per class. When analyzing performance, we assume that the same number of images is available per iris class to avoid any unwanted performance bias. We further assume that iris images are conveniently vectorized. Let $X_1^n(m), \ldots, X_N^n(m)$ be $N$ vectorized iris images of iris class $m,$ $m=1,\ldots,M,$ with superscript $n$ indicating the length of each vector.  Note that in our analysis we treat all vectors as column vectors. 

To ensure a workable model that can be used to analyze the performance of iris biometrics, we turn to an Auto-Regressive (AR) model~\cite{Shumway2017} for vectorized iris data. As a model, AR has two outstanding properties. 1) It is driven by white Gaussian noise passed as an input to a linear shift-invariant filter. Furthermore, 2) the model captures dependencies among entries in $X_i^n(m),$ that is, the entries in a vector $X^n$ are related through the following equation 
\begin{equation} 
X_t = \sum_{i=1}^p \alpha_i X_{t-i} + \eta_t, 
\label{eq:AR_model}
\end{equation}
where $\alpha_i$ is the parameter of the model, $\eta_t$ is a sample of white Gaussian noise process with mean zero and variance $\sigma^2_{\eta},$ and $p$ is the parameter that determines the order of the model. 

If AR model does not provide a reasonable fit to vectorized data, several variants of AR such as applying it to log or exponentially transformed data or to high order difference data (AR(Integrated)MA) \cite{Shumway2017} may lead to a better fit.  

Noticing that we are dealing with a linear difference equation and thus with a description of a linear system, the frequency response of the AR model in (\ref{eq:AR_model}) is easy to derive 
\begin{equation}
H(f) = \frac{1}{1+\sum_{k=1}^p \alpha_k \exp\left(-j 2 \pi f k \right)},
\label{eq:transfer_function_ARIMA}
\end{equation}
where we use $f$ to denote frequency.

Since the AR model describes the stationary behavior of data (usually time or spatial series), by knowing the transfer function of the model and the power spectrum of the driving process (which is in our case a white Gaussian noise process with zero mean and variance $\sigma^2_{\eta}$), we can write an equation of the power spectral density (PSD) of the data $X_t$ 
\begin{equation} 
S_X(f)=\sigma^2_{\eta} |H(f)|^2,
\label{eq:PSD_ARIMA}
\end{equation}
where $S_X(f)$ is the notation for the PSD on the output of the linear filter.  

Given an AR model, each iris vector $X_i^n(m)$ is a realization of the random process described by (\ref{eq:AR_model}). Since the random process is driven by a Gaussian noise process and the model is linear, the process in (\ref{eq:AR_model}) is also Gaussian. Thus, 
\begin{equation}
X_i^n(m) \sim {\cal N} ({\boldsymbol\mu}(m),\mathbf{K}(m)),
\label{eq:notation_1}
\end{equation} 
where ${\boldsymbol\mu}(m)$ is the mean (in our analysis we adjust it to be $\mathbf{0}$) and $\mathbf{K}(m)$ is the covariance matrix of the entries of the $i$-th vectorized iris image of the $m$-th class. Each iris class is fitted with a unique AR model. 
%\color{purple} Since the input noise variance $\sigma_W^2,$ parameters $\alpha_k$  and the order of AR model described by $p$ are not known in practice, they must be estimated from available data. (fix sentence) \color{black} 

%%%%%%%%%%%%%%%%%%%%%%%%%%%%%%%%%%%%%%%%%%%%%%%%%%%%%%%%%%%%%%%%%%%%%%%
\subsection{Classical Approach to the Estimation of Maximum Population}
% mathematically intractable; no closed form expressions for ......
%%%%%%%%%%%%%%%%%%%%%%%%%%%%%%%%%%%%%%%%%%%%%%%%%%%%%%%%%%%%%%%%%%%%%%%
%
Given probability models for the data of each class and the class dependencies, an optimal approach to the analysis of iris uniqueness is to state the problem of matching a query iris image $Y^n$ to one of $M$ iris classes as an $M$-ary detection problem \cite{VanTrees01}. A direct performance analysis for this problem requires forming a $(M-1)$ dimensional vector of likelihood ratios and evaluating their joint probability density under the assumption that the query data belong to one of $M$ distinct iris classes. Mathematically, performance analysis for this problem becomes quickly intractable, since the expression for the joint probability density function of the vector of likelihood ratios is not straightforward to develop. Furthermore, it is hard to implement in practice. Seeking for an alternative solution, one may turn to an analysis of $M(M-1)/2$ binary detection problems, an approach that is often used in practice. By applying the Union bound~\cite{Boole1847}, the probability of error in an M-ary problem can be upper bounded by a sum of binary error probabilities.   

Denote by $P(error)$ the average probability of error in an $M$-ary detection problem and by $P(error|H_m)$ the conditional probability of error, given that data are generated by Class $m,$ $m = 1,\ldots,M,$ we refer to it as hypothesis $H_m.$  Assuming equal prior probability for each class $m,$ the average probability of error is given as
\begin{equation} 
P(error) = \frac{1}{M} \sum_{m=1}^M P(error| H_m). 
\end{equation}
After expanding  $P(error|H_m)$ as $P(\bigcup_{k=1, \ k \neq m}^M H_k|H_m)$ and applying the Union bound, the equation above yields 
\begin{equation} 
P(error) \leq \frac{1}{M} \sum_{m=1}^M \sum_{k = 1, \ k \neq m}^M P(H_k|H_m),  
\label{eq:union_bound}
\end{equation}
where $P(H_k|H_m)$ is the error in a binary detection problem for the pair of classes $k$ and $m.$ 

The bound (\ref{eq:union_bound}) establishes a link between the total probability of recognition error and the number of iris classes $M$ and thus presents a basis for the analysis of the maximum population of iris biometrics. Despite being much simplified compared to the original $M$-ary detection problem, the bound does not yield a general explicit relationship between $P(error)$ and $M$ and becomes hard to evaluate in practice due to the complex nature of practical data. 
%the bound is a function of pairwise conditional error probabilities that must be quantified. 

To take our analysis of the maximum iris population further, in the following subsections, we will first develop an expression for the log-likelihood ratio statistic and analyze its probability distribution. Then we will return to the bound on $P(error)$ and suggest two alternative approaches that yield an explicit relationship not only between $P(error)$ and $M,$ but also involve the quality of iris data (see \cite{IREX-III} for the definitions and standards on iris quality for iris biometrics).

 %%%%%%%%%%%%%%%%%%%%%%%%%%%%%%%%
\subsection{Log-likelihood Ratio}
%%%%%%%%%%%%%%%%%%%%%%%%%%%%%%%%%
%\label{sect:model}  
Given an iris dataset composed of $M$ iris classes, with data of each class being vectorized and then fitted with an AR description,  as outlined in Sec. \ref{sec:ARMA_model}, the origin of a query vector $Y^n$ can be tested using classical detection theory approaches. Since we have a probability model for data of each class, however parameters of the models are estimated from data, we appeal to Generalized Likelihood Ratio Test (GLRT)~\cite{VanTrees01} to find which of $M$ classes is the origin of vector $Y^n.$  While our peers may find this approach outdated (too classical compared to modern deep learning-based approaches), unlike deep learning approaches, this model guarantees an insightful performance analysis, which is a powerful justification within the scope of this work.     

Given $M(M-1)/2$ pairwise binary detection problems to solve, we form a log-likelihood statistic for every pair. For testing the hypothesis ``class $m$ is the true class'' versus ``class $k$ is the true class,'' it is given as
\begin{equation}
\Lambda(m,k) = \frac{1}{N} \sum_{j=1}^N \ln \frac{f(Y^n_j| H_m )}{f(Y^n_j| H_k )},    
\label{eq:loglikelihood_statistic} 
\end{equation} 
where $f(Y^n_j| H_m)$ is the conditional pdf of the $j$-th copy of vectorized iris data $Y^n,$ conditioned on class $m.$ After involving the model in (\ref{eq:notation_1}), the log-likelihood statistic becomes
\begin{comment}
\begin{equation}
\Lambda(m,k) = -\frac{1}{2N} \sum_{j=1}^N Y^{nT}_j\left(\mathbf{K}^{-1}(m) - \mathbf{K}^{-1}(k) \right) Y^n_j - 
\frac{1}{2} \ln \det \left( \mathbf{K}(m)\mathbf{K}^{-1}(k)\right). 
\label{eq:loglikelihood_normal} 
\end{equation}
\end{comment}
\begin{multline}
\Lambda(m,k) = -\frac{1}{2N} \sum_{j=1}^N Y^{nT}_j\left(\mathbf{K}^{-1}(m) - \mathbf{K}^{-1}(k) \right) Y^n_j \\ - 
\frac{1}{2} \ln \det \left( \mathbf{K}(m)\mathbf{K}^{-1}(k)\right). 
\label{eq:loglikelihood_normal} 
\end{multline}

The test statistic $\Lambda(m,k)$ is then compared to a threshold to conclude which class ``generated'' the vector $Y^n.$  We tentatively set the value of the threshold to zero, since no prior information about the frequency of use of any two classes is available to us, and thus the binary test to perform is given as 
\begin{equation}
\Lambda(m,k)  \underset{H_m}{ \overset{H_k}{\lessgtr}} 0.
\label{eq:binary_detection}
\end{equation}
Alternatively, we can vary the value of the threshold on the right-hand side of the inequality and analyze $P(H_k| H_m)$ as a function of the threshold.   

%%%%%%%%%%%%%%%%%%%%%%%%%%%%%%%%%%%%%%%%%%%%%%%%%%%%
\subsection{Asymptotic Case of Log-likelihood Ratio}
%%%%%%%%%%%%%%%%%%%%%%%%%%%%%%%%%%%%%%%%%%%%%%%%%%%%
%
When the number of entries in a vectorized iris image is large, that is $n$ is large, (\ref{eq:loglikelihood_normal}) can be replaced by an asymptotic expression involving the power spectral density of the AR model.  It can be easily demonstrated that $\Lambda(m,k)$ in the asymptotic case can be written as  
\begin{comment}
\begin{equation}
\Lambda(m,k) = - \sum_{i=0}^{n-1} \left\{ \left( \frac{1}{S_m(f_i)} - \frac{1}{S_k(f_i)} \right) \sum_{j=1}^N \frac{|y_j(f_i)|^2}{N} + \ln \left( \frac{S_m(f_i)}{S_k(f_i)} \right) \right\}
 = - \sum_{i=0}^{n-1}\lambda(f_i),
\label{eq:loglikelihood_asymptotic_1}
\end{equation}
\end{comment}
\begin{multline}
\Lambda(m,k) = \\
- \sum_{i=0}^{n-1} \left\{ \left( \frac{1}{S_m(f_i)} - \frac{1}{S_k(f_i)} \right) \sum_{j=1}^N \frac{|y_j(f_i)|^2}{N} + \ln \left( \frac{S_m(f_i)}{S_k(f_i)} \right) \right\} \\
  = - \sum_{i=0}^{n-1}\lambda(f_i),
\label{eq:loglikelihood_asymptotic_1}
\end{multline}
where $y^n$ is the Fourier transform of $Y^n,$ $S_m(f_i)$ is the $i$-th sample of the power spectral density of the $m$-th class (for an insightful explanation of the result see p. 36 of Kay~\cite{Kay98}), and $\lambda(f_i)$ is the $i$-th component of the log-likelihood ratio statistic.

%%%%%%%%%%%%%%%%%%%%%%%%%%%%%%%%%%%%%%%%%%%%%%%%%%%%%
\subsection{Analysis of Error Probability, continued}
\label{sec:analysis_of_error_probability}
%%%%%%%%%%%%%%%%%%%%%%%%%%%%%%%%%%%%%%%%%%%%%%%%%%%%%
%
%
Given a binary detection problem involving two classes, $Class \ m$ and $Class \ k,$ an error will occur in two cases: Case 1: $Y^n$ originated from $Class \ m,$ but $\Lambda(m,k) < 0;$ and Case 2: $Y^n$ originated from $Class \ k,$ but $\Lambda(m,k) > 0.$  The first case describes $P(H_k|H_m),$ while the second case describes $P(H_m| H_k).$  Both conditional probabilities of error can be expressed in terms of the conditional probability density function of $\Lambda(m,k),$ assuming one or the other class is the true class. 

Consider $P(H_k | H_m) = P(\Lambda(m,k) < 0|H_m).$  In (\ref{eq:loglikelihood_asymptotic_1}), random vector $y^n$ is complex Gaussian under either hypothesis, since $y^n$ is a linear transformation of a Gaussian vector. To find the conditional probability of error $P(H_k| H_m),$ we need a closed form expression for the conditional probability density function (p.d.f.) of $\Lambda(m,k)$ under $H_m.$  

Assuming that $y^n$ is from Class $m$ implies that $y(f_i) \sim {\cal CN}(0,S_m(f_i)),$  where ${\cal CN}$ denotes ``complex normal,''   
\[ 
 y(f_i)\left(\frac{1}{S_m(f_i)}-\frac{1}{S_k(f_i)}\right)^{1/2} \]
\[
\sim {\cal CN}\left(0,S_m(f_i)\left(\frac{1}{S_m(f_i)}-\frac{1}{S_k(f_i)}\right)\right)
\]   
and 
\[ 
\lambda(f_i) =   \left( \frac{1}{S_m(f_i)} - \frac{1}{S_k(f_i)}\right) \sum_{j=1}^N\frac{|y_j(f_i)|^2}{N} + \ln \left( \frac{S_m(f_i)}{S_k(f_i)} \right) 
\]
is a $N$-Erlang random variable with the p.d.f. 
\begin{equation} 
f_{\lambda(f_i)}(x) = \frac{N^N}{(\sigma^2_i)^N} \frac{(x-a_i)^{N-1}}{(N-1)!} \exp \left\{  -\frac{N(x-a_i)}{\sigma^2_i} \right\},  \ \ x>a_i
\end{equation} 
where $a_i = \ln (S_m(f_i)/S_k(f_i))$ and $\sigma^2_i = 1 - S_m(f_i)/S_k(f_i).$

The entries $\lambda(f_i)$ in the test statistic $\Lambda(m,k)$ are independent, but not identically distributed. Therefore, a closed-form expression for the conditional pdf of $\Lambda(m,k)$, assuming that the data are generated by $Class \ m$, is not straightforward to find. At this point, we can take our analysis further by involving the Chernoff bound \cite{VanTrees01} on $P(H_k|H_m).$ Instead, equipped with the form of the p.d.f. for $\lambda(f_i),$ the well developed theory of error correction codes \cite{Cover2006, VanLint1999}, and a deep insight into Daugman's analysis of IrisCode \cite{Daugman2016, Daugman2021}, we reverse the direction of our analysis. In the following two subsections, we analyze the uniqueness of iris biometrics from the perspective of the sphere packing argument \cite{Cover2006} and by developing a Daugman-like bound \cite{Daugman2016}. Both provide an explicit relationship of $P(error)$ on the number of classes $M$ and an average quality of iris data in a considered iris dataset.     
%\color{purple} (Rephrase since we are not using Chernoff Bound) An alternative solution is to find the moment generating function of $\Lambda(m,k)$ first, then develop a tight exponential upper bound on the conditional probability $P(H_k|H_m).$ We involve the Chernoff bound. \color{black}
%take the inverse Laplace transform of the expression. 
%to obtain the conditional pdf of $\tilde{\Lambda}(m,k).$  

%%%%%%%%%%%%%%%%%%%%%%%%%%%%%%%%%%%%%%%%%%%%%%%%%%%%%%%%%%%%%%%%%%%%%%%%%%%%%
%\subsection{Fitting chi-square pdf into a histogram of relative entropy values}
\subsection{Analysis of iris uniqueness using sphere packing argument}
\label{sec:Sphere_Packing_Theory}
%%%%%%%%%%%%%%%%%%%%%%%%%%%%%%%%%%%%%%%%%%%%%%%%%%%%%%%%%%%%%%%%%%%%%%%%%%%%%
%
As justified in Sec. \ref{sec:analysis_of_error_probability}, the log-likelihood ratio test statistic is a sum of weighted exponential random variables. While no method for direct evaluation of its pdf is known, 
%Given pairwise log-likelihood ratio statistics (can also be called estimates of relative entropy between distributions of two classes), 
a plot of the relative frequency of the log-likelihood statistic can be approximated by a chi-square p.d.f. formed by adding $K$ iid squared complex Gaussian random variables each with zero mean and variance $P.$ The parameter $K$ is the number of degrees of freedom of the fitted chi-square p.d.f. Since $K$ and $P$ are unknown, they must be estimated from empirical data.  

The fitted chi-square p.d.f. allows us to interpret the problem of finding the maximum iris population as a Gaussian sphere packing result. Suppose an encoding strategy is available to map ideal iris images (iris images with no noise or distortions) of $M$ distinct iris classes into unique Gaussian codewords, each of length $K.$ Each codeword is drawn iid from a Gaussian distribution with zero mean and variance $P.$ Suppose further that an iris image of one of the $M$ classes (for example, of Class $m$) submitted for authentication or recognition is modeled as a noisy version of the ideal codeword of Class $m.$ The noise is zero mean Gaussian with variance $N$ in each of $K$ dimensions. Thus, for a given Class $m$ the iris image submitted for authentication is mapped into a point within a $K$-dimensional sphere with radius $\sqrt{K N}$ around the codeword of Class $m.$ Since the Gaussian sphere containing codewords of $M$ classes has radius $\sqrt{K(P+N)},$  the maximum number of classes, assuming that the distortion of iris images submitted for authentication is bounded, can be obtained by dividing the volume of a $K$ dimensional sphere containing all codewords by the volume of the small sphere representing noise in the data of particular iris class. Thus, 
\begin{equation}
M \leq \left( 1 + \frac{P}{N} \right)^{K/2}. 
\label{eq:SpherePackingBound}
\end{equation}
See \cite{Cover2006} for a more insightful description. 

% Moved from the Implementation section 
\begin{figure*}[!t]
\centering
    \includegraphics[width=6in]{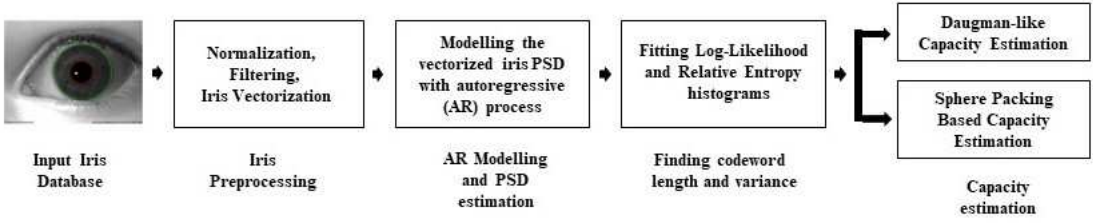}
    \caption{An overall block diagram summarizing the proposed methodology to finding the maximum population of an iris database.
    }
    \label{fig:Pipeline}
\end{figure*} 

%%%%%%%%%%%%%%%%%%%%%%%%%%%%%%%%%%%%%%%%%%%%%%%%%%%%%%%%%%%%%%%%%%%%%%%
\subsection{A Daugman-like approach to the analysis of iris uniqueness}
\label{sec:Daugman_Bound_Theory}
%%%%%%%%%%%%%%%%%%%%%%%%%%%%%%%%%%%%%%%%%%%%%%%%%%%%%%%%%%%%%%%%%%%%%%%
%
Similar to the sphere packing argument presented in the previous section, Daugman-like analysis of iris uniqueness is based on the assumption that data of iris classes are mapped into a space in which each iris class is presented by an independent Gaussian codeword of length $K$ with zero mean and variance $P,$ where $K$ and $P$ are defined above. This mapping ensures that the asymptotic pairwise log-likelihood ratios (here interpreted as a distance between two codewords) are independent chi-square distributed random variables with $K$ degrees of freedom.  The asymptotic log-likelihood ratio can be also replaced with an estimate of the relative entropy between the pdfs of two iris classes. This leads to a new interpretation of the  distance measure as a means to also measure the quality of iris data. Its introduction allows a rate-distortion interpretation of the problem of finding the maximum iris population that an iris recognition system can sustain, similar to how error-correction bounds in coding theory relate the maximum population of binary code to the minimum Hamming distance between codewords~\cite{VanLint1999}.  To be specific, the introduction of such a metric will lead to a new performance bound that relates the size of the iris population covered by the recognition algorithm and the quality of iris biometric data while ensuring a small probability of recognition error.     

At this point of our analysis, in addition to the asymptotic log-likelihood ratio statistic we introduce the relative entropy between the probability density functions of two classes $m$ and $k.$  The relative entropy is defined as the expected value of the log-likelihood ratio in %(\ref{eq:loglikelihood_normal}) and 
(\ref{eq:loglikelihood_asymptotic_1}) 
\begin{equation}
d(m,k) = \mathbb{E} \left[ \Lambda(m,k) \right] = \sum_{i=0}^{n-1} \left\{ \frac{S_m(f_i)}{S_k(f_i)} - \ln \frac{S_m(f_i)}{S_k(f_i)} - 1 \right\},  
\label{eq:relative_entropy}
\end{equation} 
where $\mathbb{E}$ is the notation for the expected value operator. Since the power spectral densities of different iris classes are not known to us, they are first estimated from available class data and then plugged in the expression for the relative entropy in place of the true unknown power spectral densities.  

With estimated relative entropy as a distance metric, the bound on the maximum population of enrolled iris population is straightforward to develop. We follow an argument similar to Daugman's that the imposter distance between a pair of distinct iris classes can be fitted with a chi-square pdf with $K$ degrees of freedom. 
%This allows us to interpret the distance (relative entropy) between any two iris classes as having $K$ independent identically distributed chi-square components, each with $1/2$ degree of freedom. 
%
Then the error to enroll can be mathematically described as 
\begin{equation} 
P(error\ to\ enroll) = 1 - P\left( \bigcap_{m=1}^M d(m,M+1) > \tau\right),
\label{eq:error_to_enroll}
\end{equation} 
where $d(m,M+1)$ is the distance between a previously successfully enrolled class $m$  and a new (not yet enrolled) class $M+1$ and $\tau$ is a minimum distance between two codewords for them to represent two distinct classes.   
%pre two iris classes interpreted as a chi-square with $N$ degrees of freedom.  
Since pairwise distances between iris classes are independent identically distributed chi-square random variables, (\ref{eq:error_to_enroll}) can be rewritten as 
%with $K$ degrees of freedom The histogram and fitted pdf will be illustrated in Section ?????. Having chi-square pdf fitted into the histogram of $d(m,k)$ with $k \neq m,$  ..... 
\begin{equation}
P(error\ to\ enroll) = 1 - \left\{ 1 - P(d(m,M+1) \leq \tau ) \right\}^M \leq \delta,  \label{eq:error_to_enroll_v2}  
\end{equation}
%\color{blue} where the first equality is due to the assumption that codewords for each class can be selected as independent and such that the pairwise distances are independent as well. \color{black} 
Inverting the inequality for $M$ results in 
\begin{equation} 
M \leq \frac{\log (1 - \delta)}{\log \left\{ 1 - FMR(\tau) \right\}},
\label{eq:Max_Pop_Bound_Daugman}
\end{equation}
 where $P(d(m,M+1) \leq \tau)$ is replaced with $FMR(\tau),$ abbreviation for False Match Rate as a function of the distance between two codewords $\tau.$

%%%%%%%%%%%%%%%%%%%%%%%%%%%%%%%%%%%%%%%%%
\section{Illustration of the Methodology}
%\section{Implementation}
\label{sec:implementation}
%%%%%%%%%%%%%%%%%%%%%%%%%%%%%%%%%%%%%%%%%
%
The following section provides a basic illustration of the above-mentioned theory on two small subsets of CASIA-IrisV3 Interval and BATH datasets. As the sphere packing bound (a purely theoretical result) and a methodology to approach it are the main focus of this work, we do not intend to extend their illustration beyond this example to avoid obscuring the theory and methodology.

The power spectral densities for each dataset are estimated using the AR modeling described in Section \ref{sec:ARMA_model}, along with justification for selecting the correct model order. The estimated power spectral densities for each class are then substituted into the distance metrics in (\ref{eq:loglikelihood_asymptotic_1}) and (\ref{eq:relative_entropy}). For both metrics, the histograms are fitted using a chi-square distribution, as outlined in Section \ref{sec:Sphere_Packing_Theory}, and the resulting degrees of freedom and variance are substituted into (\ref{eq:SpherePackingBound}) and (\ref{eq:Max_Pop_Bound_Daugman}) to find the maximum population based on iris quality and recognition error. Figure \ref{fig:Pipeline} shows an overview of the methodology presented and discussed in subsequent sections. 

%\color{red} 
%This block diagram must be replaced. ``Masek's segmentation'' must be removed or replaced with ``iris segmentation,'' ``ZigZag Iris Vectorization'' must be replaced with ``Iris Vectorization.'' We may keep everything else.  
%\color{black}

%
%\begin{figure}[!t]
%\centering
%\begin{subfigure}{.5\textwidth}
%  \centering
%  \includegraphics[scale = 0.25]{Graphics/CASIAOccludedImageS1001L09.jpg}
%  \caption{Example excluded image from CASIA dataset.}
%  \label{fig:sub1_Occluded}
%\end{subfigure}%
%\\
%\begin{subfigure}{0.5\textwidth}
%  \centering
%  \includegraphics[scale = 0.25]{Graphics/BATHOccludedImage8L.jpg}
%  \caption{Example excluded image from BATH dataset.}
%  \label{fig:sub2_Occluded}
%\end{subfigure}
%\caption{Example iris images from CASIA (\ref{fig:sub1_Occluded}) and BATH (\ref{fig:sub2_Occluded}). \ref{fig:sub1_Occluded} shows both upper occlusion, from eyelashes, and lower occlusion, from eyelid. \ref{fig:sub2_Occluded} shows dominate eyelash occlusion. }
%\label{fig:Occluded_Data}
%\end{figure}
% 

%\begin{figure*}[!t]
%\centering
%    \includegraphics[scale = 0.65]{Graphics/PreProcessingStepsFigure.JPG}
%    \caption{Example segmentation and pre-processing steps of a single BATH iris image before ZigZag vectorization is performed. 
%    }
%    \label{fig:PreProcessingPipeline}
%\end{figure*} 

%%%%%%%%%%%%%%%%%
\subsection{Data}
%%%%%%%%%%%%%%%%% 
% 
The following illustration is carried out on two small datasets: Chinese Academy of Sciences' Institute of Automation (CASIA) CASIA-IrisV3 Interval \cite{CASIADatabase} and University of Bath (BATH) Iris Image Database \cite{Monro2007}. CASIA-IrisV3 Interval contains $2,639$ near-infrared (NIR) illuminated images, each having a resolution of $320 \times 280$ pixels, and a total of $249$ subjects. A trial subset of the BATH dataset 
%contains $32,000$ NIR iris images with $800$ subjects, however a smaller portion of this dataset was used (as the larger dataset is no longer publicly available). Therefore, the smaller `sample' BATH dataset that is used 
contains $1,000$ images, each with resolution $960 \times 1280$ pixels, and $25$ subjects (with each subject having $20$ images for both the left and right eye). These datasets are chosen due to their high-quality iris images that show the rich texture around the pupil. 

%To ensure that we are analyzing iris uniqueness correctly, 
Some data reduction is performed on each iris dataset to balance the dataset and extract the highest-quality iris images. The CASIA-IrisV3 Interval dataset is reduced by removing images with more than $50\%$ iris occlusion and then excluding classes with fewer than $10$ images per class resulting in only $21$ remaining classes with a total of $210$ images. 
%From the remaining, classes were removed that had more than $50\%$ occlusions present, such as eyelashes or upper/lower eyelid. Through this simple elimination procedure, only $21$ classes remained for CASIA with a total of $210$ images. 
After a similar analysis of images in the BATH dataset, $40$ iris classes were retained with a total of $800$ images. 
%the first step was ignored, since each iris class contained a balanced number of $20$ images, and the high occlusion classes were omitted. The remaining classes for the BATH dataset are $40$ with a total $800$ images. Figure \ref{fig:Occluded_Data} shows an example of an occluded class from each dataset.
These smaller sets of data are used for the remainder of the paper.

%%%%%%%%%%%%%%%%%%%%%%%%%%%%%%%%%%%%%%%%%%%%
\subsection{Segmentation and Pre-Processing}
%%%%%%%%%%%%%%%%%%%%%%%%%%%%%%%%%%%%%%%%%%%%
%
%To begin the analysis of the uniqueness, 
Iris images are segmented, normalized, and Gabor-filtered. Since a majority of the texture of the iris is located close to the pupil, only half of the filtered image is considered and the remainder is discarded. 
%Figure \ref{fig:PreProcessingPipeline} shows the processing of a single iris. 
Once preprocessed, the complex-valued image is unwrapped into a single vector (vectorized). We considered multiple methods for vectorization and concluded with Zigzag vectorization \cite{Gonzalez2008}. It unwraps the real-valued portion of the image into a one-dimensional vector by applying a diagonal scan from the top left corner of the image to its bottom right corner. The same unwrapping is applied to the imaginary-valued portion of the image, and then the real and imaginary valued one-dimensional vectors are concatenated into a long data vector (for our data size at $4,800$ pixels). The applied vectorization may not be the best existing method to vectorize iris images, however, it is suitable enough to illustrate the proposed theory and bounds.  
%This ZigZag vectorization method was adapted to persevere the spatial correlation of the iris patterns while transitioning the 2D image to a 1D representation. ZigZag was introduced in \cite{Gonzalez2008} and previously effectively used in combination with Discrete Cosine Transform to accomplish lossy compression. It was also adopted in application to biometrics as fingerprint \cite{Wang2010} and hand (palm print) biometrics \cite{Ergen2016} to unwrap 2D DCT filtered images to a 1D vector. 
For our application, Zigzag helps in reducing the variance in AR coefficients and eliminating induced periodicity in the estimated power spectral density incorporated due to spatial distortion of the iris patterns in horizontal (row-based) or vertical (column-based) vectorization.  
%
%\begin{figure}[!t]
%\centering
%    \includegraphics[width=3in]{Graphics/ZigZag.png}
%    \caption{An illustration of the ZigZag vectorization of iris images as adapted in this work}
%    \label{fig:ZigZag}
% \end{figure}

%%%%%%%%%%%%%%%%%%%%%%%%%%%%%%%%%%%%%%%%
\subsection{Estimation of Power Spectra} 
%%%%%%%%%%%%%%%%%%%%%%%%%%%%%%%%%%%%%%%%
%
As stated in Sec. \ref{sec:ARMA_model}, to ensure a workable model that can be used to analyze the performance of iris biometrics, we turn to an Auto-Regressive (AR) model for the vectorized iris data. The analysis of maximum population is based on the successful implementation of (\ref{eq:loglikelihood_asymptotic_1}) and (\ref{eq:relative_entropy}) which in turn rely upon estimates of the power spectral densities obtained from data of iris classes. These estimates are obtained through (i) finding the optimal order for the AR model given our iris data, \ref{sec:Optimal_Order} and (ii) using Burg's Maximum Entropy Method \cite{Burg1975} to find high-quality spectral estimates for each iris class, \ref{sec:AR_Implementation}.

%%%%%%%%%%%%%%%%%%%%%%%%%%%%%%%%%%%%%%%%%%%
\subsubsection{Finding Optimal Model Order}
\label{sec:Optimal_Order}
%%%%%%%%%%%%%%%%%%%%%%%%%%%%%%%%%%%%%%%%%%%
%
Estimating the appropriate model order is essential in the performance of the AR model. Having a large order ensures a better fit into data, however it also increases the complexity of the implementation and can lead to fitting the AR model to noise rather than to signal. To find the optimal model order of estimated power spectra we involve the Akaike Information Criterion (AIC) in conjunction with the AR method in \textsc{Matlab}. 
%to find the optimal model order of estimated power spectra.
%
%\begin{figure*}[!t]
%\centering
%    \includegraphics[scale = 0.45]{Graphics/Textured_Irises_CASIA.JPG}
%    \caption{Subset of CASIA-IrisV3 Interval dataset that is used to see if the texture had an effect on optimal AR model order. Subjects are chosen by varying texture, with Subject 199R having a very fine texture to Subject 104L having a very rough texture. 
%    }
%    \label{fig:TexturedIrisesTable}
%\end{figure*} 
%
\begin{figure}[!t]
\centering
\begin{subfigure}{0.35\textwidth}
  \centering
  \includegraphics[width = \textwidth]{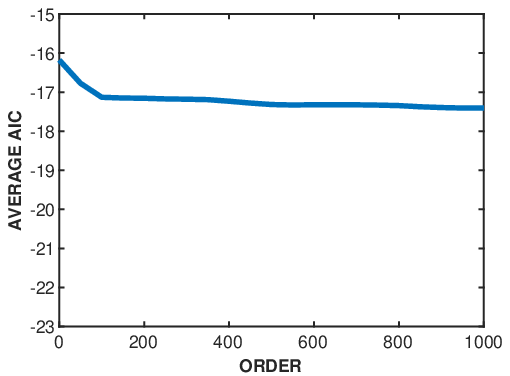}
  \caption{CASIA-IrisV3 Interval}
  \label{fig:AIC_CASIA}
\end{subfigure}%
\\
\begin{subfigure}{0.35\textwidth}
  \centering
  \includegraphics[width = \textwidth]{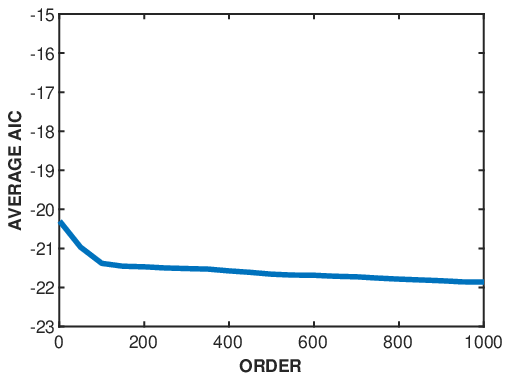}
  \caption{BATH}
  \label{fig:AIC_BATH}
\end{subfigure}
\caption{Plots of average AIC values for a single class in (a) CASIA-IrisV3 Interval  and (b) BATH with optimal Gabor filter parameters of $f_0 = 1/9$ and $\sigma = 0.5.$ (AIC scores were found for each iris image separately, then the average scores were plotted for analysis.)}
\label{fig:AIC_Curves}
\end{figure}

To begin, a subset of iris classes were extracted from each dataset based on varying texture levels, from very fine to rough texture, to see if the structure of the iris affects model order. Through extensive search, the model order is varied along with two parameters of Gabor filters, the center frequency, $f_0,$ and the filter bandwidth, $\sigma.$ The goodness of fit of each estimated model is measured using the value of AIC for each tested model order. This is repeated for each class in the subset on each iris sample until the optimal order, center frequency, and bandwidth are found in which the AIC converges. Figure \ref{fig:AIC_Curves} shows the AIC curves for CASIA-IrisV3 Interval and BATH with the selected order at 100 and Gabor filter parameters of $f_0 = 1/9$ and $\sigma = 0.5.$ Although Figure \ref{fig:AIC_Curves} only shows one iris class for each dataset, we numerically confirmed that  each class can be parameterized by the same model order. Choosing a higher order leads to the same performance as an order of 100.

%%%%%%%%%%%%%%%%%%%%%%%%%%%%%%%%%%%%
\subsubsection{AR Implementation}
\begin{figure}[!t]
\centering
\begin{subfigure}{0.35\textwidth}
  \centering
  \includegraphics[width=\textwidth]{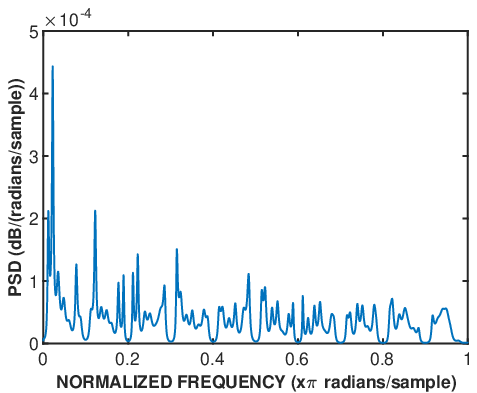}
  \caption{CASIA IrisV3 Interval}
  \label{fig:EstPSDCasia}
\end{subfigure}%
\\
\begin{subfigure}{0.35\textwidth}
  \centering
  \includegraphics[width=\textwidth]{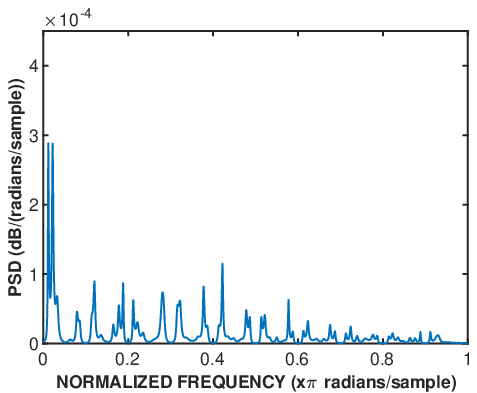}
  \caption{BATH}
  \label{fig:EstPSDbath}
\end{subfigure}
\caption{Estimated Power Spectral Densities for both datasets through the use of \textsc{Matlab's} \textit{pburg} method.}
\label{fig:EstimatedPSDsPBURG}
\end{figure}

%\subsubsection{ARMA Implementation}
\label{sec:AR_Implementation}
%%%%%%%%%%%%%%%%%%%%%%%%%%%%%%%%%%%%
%
Burg's Maximum Entropy Method, which is popular for stationary or piece-wise stationary data, 
is implemented to find high-quality spectral estimates for each iris class. For both databases, the power spectral densities (PSDs) were estimated for each image in each class through the use of \textsc{Matlab}'s Signal Processing Toolbox function \textit{pburg}, with the found optimal order from \ref{sec:Optimal_Order} as the model order input. The vectorized iris images are found to be low-frequency signals, as shown in Figure \ref{fig:EstimatedPSDsPBURG}. 

After the completion of Burg's Method, a selection of PSDs is averaged to create a high-quality enrollment spectrum for each class, and the remaining spectra are used for the authentication process. The number of samples used for the enrollment spectrum depends on the available amount of iris images per class, with BATH having 20 images available per iris class and CASIA-IrisV3 Interval having a maximum of 10 images. With this in mind, half (50\%) of the class's iris images are used to find the enrollment spectrum, while the remaining percentage is used for authentication. These estimates are used in (\ref{eq:loglikelihood_asymptotic_1}) and (\ref{eq:relative_entropy}) to empirically find the maximum population given the sphere packing bound and Daugman-like bound, shown in \ref{sec:SpherePackingBoundImplementation} and \ref{sec:DaugmanLikeBoundImplementation}.

%%%%%%%%%%%%%%%%%%%%%%%%%%%%%%%%%%%%%%%%%%
\subsection{Fitting Imposter Distribution}
\label{sec:FittingImposters}
\begin{figure}[!t]
\centering
\begin{subfigure}{0.35\textwidth}
  \centering
  \includegraphics[width=\textwidth]{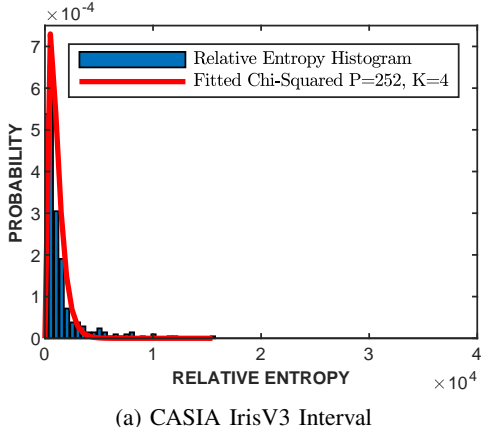}
  \caption{CASIA IrisV3 Interval}
  \label{fig:FitChi_CASIA}
\end{subfigure}
\\
\begin{subfigure}{0.35\textwidth}
  \centering
  \includegraphics[width=\textwidth]{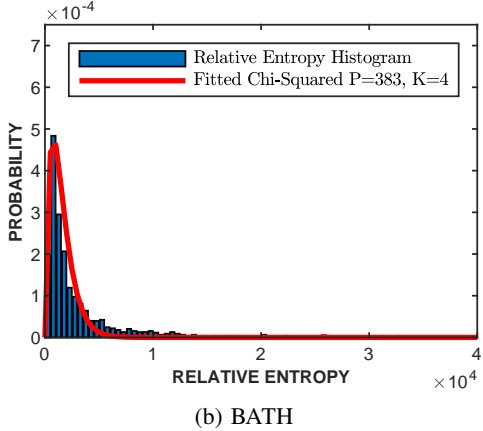}
  \caption{BATH}
  \label{fig:FitChi_BATH}
\end{subfigure}
\caption{Relative Entropy Imposter distributions for both datasets with best-of-fit chi-square distributions. CASIA-IrisV3 Interval having $K=4$ degrees of freedom and a fitted variance of $P=252$, shown in (a), and BATH having $K=4$ degrees of freedom and a fitted variance of $P=383$, shown in (b). }
\label{fig:FitChiCurves}
\end{figure}

\begin{figure}[!t]
\centering
\begin{subfigure}{0.38\textwidth}
  \centering
  \includegraphics[width=\textwidth]{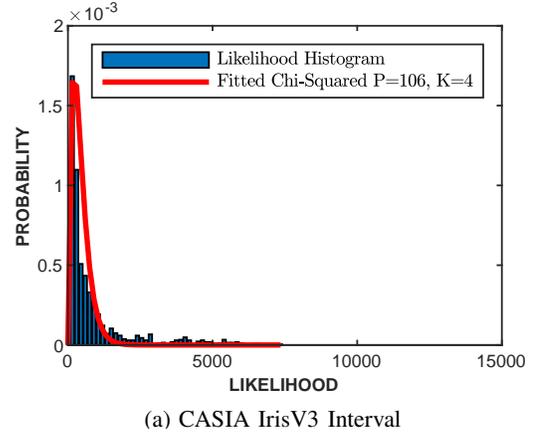}
  \caption{CASIA IrisV3 Interval}
  \label{fig:FitChi_CASIA_Likelihoods}
\end{subfigure}
\\
\begin{subfigure}{0.38\textwidth}
  \centering
  \includegraphics[width=\textwidth]{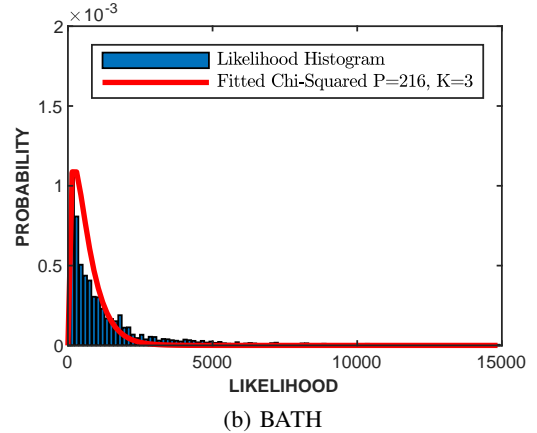}
  \caption{BATH}
  \label{fig:FitChi_BATH_Likelihoods}
\end{subfigure}
\caption{Likelihood Imposter distributions for both datasets with best-of-fit chi-square distributions. CASIA-IrisV3 Interval having $K=4$ degrees of freedom and a fitted variance of $P=106$, shown in (a), and BATH having $K=3$ degrees of freedom and a fitted variance of $P=216$, shown in (b). }
\label{fig:FitChiCurves_Likelihoods}
\end{figure}

%%%%%%%%%%%%%%%%%%%%%%%%%%%%%%%%%%%%%%%%%%
%
To find the empirical values for $K$ and $P$ from \ref{sec:Sphere_Packing_Theory} and \ref{sec:Daugman_Bound_Theory}, the imposter distributions of both likelihood values and relative entropies have to be fitted with chi-square distributions. With the use of the estimated PSDs from \ref{sec:AR_Implementation}, the pair-wise likelihood scores and relative entropies are found for all possible combinations, where $S_m$ and $S_k$ are from different classes. The best-fit chi-square distribution is found by performing an exhaustive search on the imposter histograms and finding the variance and degrees of freedom that produce the minimum least square error.  Figures \ref{fig:FitChiCurves} and \ref{fig:FitChiCurves_Likelihoods} show the distributions for both CASIA-IrisV3 Interval and BATH datasets. Both datasets have the best fit with four degrees of freedom, $K=4$, and different variances of $P_{CASIA} = 252$ and $P_{BATH} = 383$ for the relative entropy imposter distributions, shown in Figure \ref{fig:FitChiCurves}.  Figure \ref{fig:FitChiCurves_Likelihoods} shows the fitted likelihood distributions with CASIA-IrisV3 Interval having four degrees of freedom, $K=4$ and a variance of $P_{CASIA} = 106$, and BATH having three degrees of freedom, $K=3$, and a variance of $P_{BATH} = 216.$

%%%%%%%%%%%%%%%%%%%%%%%%%%%%%%%%%%%%%%%%%%%%
\subsection{Sphere Packing Bound}
\label{sec:SpherePackingBoundImplementation}
\begin{figure}[!t]
\centering
\begin{subfigure}{.35\textwidth}
  \centering
  \includegraphics[width=\textwidth]{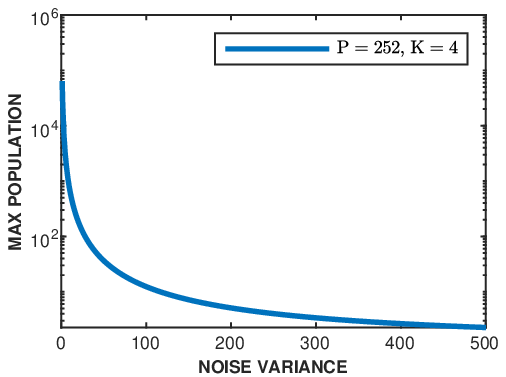}
  \caption{CASIA_IrisV3 Interval}
  \label{fig:SPB_CASIA_RE}
\end{subfigure}%
\\
\begin{subfigure}{0.35\textwidth}
  \centering
  \includegraphics[width=\textwidth]{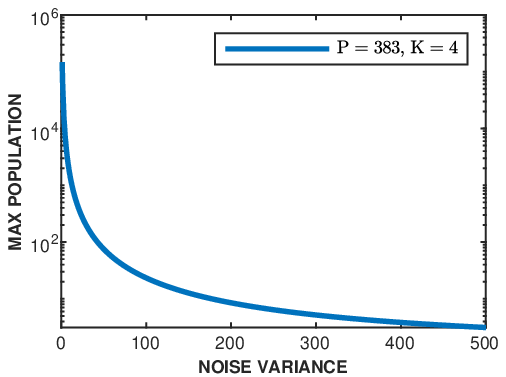}
  \caption{BATH}
  \label{fig:SPB_BATH_RE}
\end{subfigure}
\caption{Sphere Packing Bound for (a) CASIA-IrisV3 Interval and (b) BATH using the relative entropy metric.}
\label{fig:SpherePackingBounds_RE}
\end{figure}

\begin{figure}[!t]
\centering
\begin{subfigure}{.35\textwidth}
  \centering
  \includegraphics[width=\textwidth]{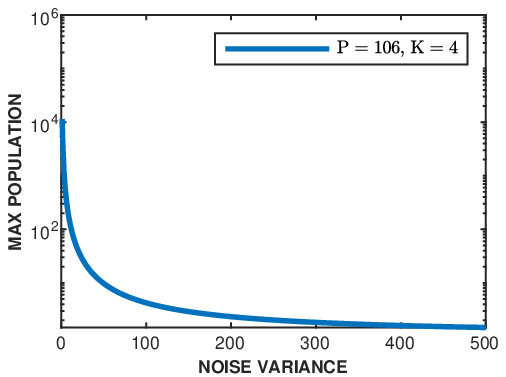}
  \caption{CASIA_IrisV3 Interval}
  \label{fig:SPB_CASIA}
\end{subfigure}%
\\
\begin{subfigure}{0.35\textwidth}
  \centering
  \includegraphics[width=\textwidth]{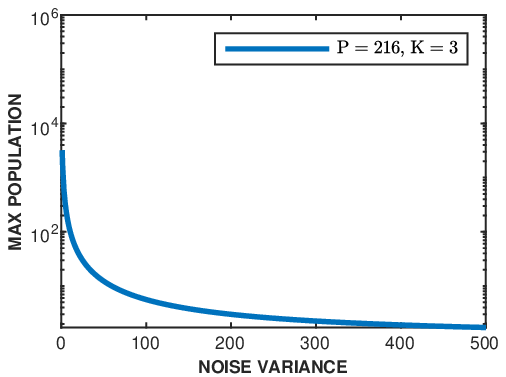}
  \caption{BATH}
  \label{fig:SPB_BATH}
\end{subfigure}
\caption{Sphere Packing Bound for (a) CASIA-IrisV3 Interval and (b) BATH using the log-likelihood metric.}
\label{fig:SpherePackingBounds}
\end{figure}

%%%%%%%%%%%%%%%%%%%%%%%%%%%%%%%%%%%%%%%%%%%%
%
Now that we have obtained the best-of-fit chi-square distributions for the relative entropy and likelihood imposter histograms, the maximum population can be empirically found through the use of the sphere packing bound presented in \ref{sec:Sphere_Packing_Theory}. Looking first at likelihoods, the found degrees of freedom and variance are used in (\ref{eq:SpherePackingBound}). For the measure of noise variance, $N,$ this depends on the quality of the iris datasets themselves. Since there is no simple method to find the noise present in the iris classes themselves, the noise variance is varied to reflect possible values (from little to extremely noisy images). Figure \ref{fig:SpherePackingBounds_RE} shows the resulting bound on the supported maximum population for CASIA-IrisV3 Interval (\ref{fig:SPB_CASIA_RE}) and BATH (\ref{fig:SPB_BATH_RE}) datasets dependent on the given noise variances with using relative entropy as a distance metric. The same procedure is implemented on the found log-likelihood fitted chi-square distributions for each dataset, and the resulting bounds are shown in Figure \ref{fig:SpherePackingBounds}. Table \ref{SpherePackingBound} shows the maximum population of each dataset given a noise variance. 

Given that the noise variance is dependently related to the quality of the iris images contained in the dataset (e.g. motion blur, focus, distance, noise, etc.), the higher quality of iris acquisition the smaller the noise variance and vice versa. This aligns with the results in which we are seeing in Table \ref{SpherePackingBound}. We can conclude that as the noise variance increases (image quality decreases) the maximum population we can support in both datasets diminishes. Since the BATH dataset contains higher-quality images than CASIA-IrisV3 Interval, the maximum supported population given the lowest noise variance ($N=1$) is $1.48 \times 10^5$ classes for relative entropy and $3.2 \times 10^3$ classes for log-likelihoods, while CASIA's is $6.40 \times 10^4$ classes for relative entropy and $1.15 \times 10^4$ classes for log-likelihoods. The reason we are seeing a higher supported population for CASIA, given sum-log-likelihoods, is due to the higher degree of freedom in the fitted histogram distribution.

%\color{purple} Should we include likelihoods for Sphere Packing Bound? \color{black}
%

\begin{table*}[h!]
\caption{Subset of Sphere Packing Bound Values given certain noise variance (N) from Figure \ref{fig:SpherePackingBounds_RE} and Figure \ref{fig:SpherePackingBounds}.}
\begin{center}
\begin{tabular}{||c || c | c|| c | c||} 
\hline
 & \multicolumn{2}{|c||}{Relative Entropy}  & \multicolumn{2}{|c||}{Likelihoods} \\
 \hline
 $Noise \ Variance \ (N)$ & $M_{CASIA}$ & $M_{BATH}$ & $M_{CASIA}$ & $M_{BATH}$ \\ [0.5ex] 
 \hline\hline
 1 & $6.40 \times 10^{4}$ & $1.48 \times 10^{5}$ & $1.15 \times 10^{4}$ & $3.2 \times 10^{3}$ \\ 
 \hline
 10 & 686 & $1.54 \times 10^{3}$ & 134 & 107 \\
 \hline
 50 & 36 & 74 & 9 & 12\\
 \hline
 100 & 12 & 23 & 4 & 5\\
 \hline
 200 & 5 & 8 & 2 & 2\\
 \hline
 300 & 3 & 5 & 2 & 2\\
 \hline
 400 & 2 & 3 & 2 & 2\\
 \hline
 500 & 2 & 3 & 2 & 2 \\ [1ex] 
 \hline
\end{tabular}
\end{center}
\label{SpherePackingBound}
\end{table*}

%%%%%%%%%%%%%%%%%%%%%%%%%%%%%%%%%%%%%%%%%%
\subsection{Daugman like bound}
\label{sec:DaugmanLikeBoundImplementation}
%%%%%%%%%%%%%%%%%%%%%%%%%%%%%%%%%%%%%%%%%%
%
%
\begin{figure}[!t]
\centering
\begin{subfigure}{.35\textwidth}
  \centering
  \includegraphics[width=\textwidth]{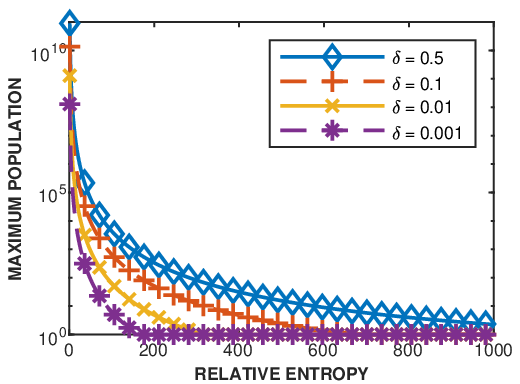}
  \caption{Bound for CASIA}
  \label{fig:SPB_CASIA}
\end{subfigure}%
\\
\begin{subfigure}{0.35\textwidth}
  \centering
  \includegraphics[width=\textwidth]{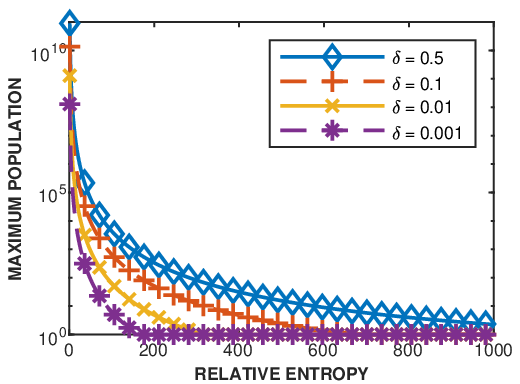}
  \caption{Bound for BATH}
  \label{fig:SPB_BATH}
\end{subfigure}
\caption{Daugman-like Bound for (a) CASIA and (b) BATH datasets.}
\label{fig:CASIA_BATH_Daugman}
\end{figure}

\begin{comment} 

\begin{figure}[!t]
\centering
    \includegraphics[width=0.35\textwidth]{Graphics/CASIA_Daugman_New_SemiLog.eps}
    \caption{Daugman-like Bound for CASIA dataset. 
    }
    \label{fig:CASIA_Daugman}
\end{figure} 
%
%
\begin{figure}
\centering
    \includegraphics[width=0.35\textwidth]{Graphics/BATH_Daugman_New_SemiLog.eps}
    \caption{Daugman-like Bound for BATH dataset. 
    }
    \label{fig:BATH_Daugman}
\end{figure} 

\end{comment} 

\begin{table*}[h!]
\caption{Subset of Daugman-Like Bound Values given a certain Relative Entropy  from Figure \ref{fig:CASIA_BATH_Daugman}.}
\begin{center}
\small\addtolength{\tabcolsep}{-5pt}
\begin{tabular}{||c || c | c||c|c||c|c||c|c|c||} 
 \hline
   & \multicolumn{2}{c||}{$\delta = 0.5$} & \multicolumn{2}{|c||}{$\delta = 0.1$} & \multicolumn{2}{|c||}{$\delta = 0.01$} & \multicolumn{2}{|c|}{$\delta = 0.001$} \\ 
  \hline
  RE & \small $M_{CASIA}$ & \small $M_{BATH}$ & \small $M_{CASIA}$ & \small $M_{BATH}$ & \small $M_{CASIA}$ & \small $M_{BATH}$ & \small $M_{CASIA}$ & \small $M_{BATH}$\\  [0.5ex] 
 \hline\hline
  1 & $1.68 \times 10^{10}$ & $8.97 \times 10^{10}$ & $2.56 \times 10^{9}$ & $1.36 \times 10^{10}$ & $2.44 \times 10^{8}$ & $1.33 \times 10^{9}$ & $2.43 \times 10^{7}$ & $1.30 \times 10^{8}$ \\ 
 \hline
  10 & $5.73 \times 10^{6}$ & $3.02 \times 10^{7}$ & $8.71 \times 10^{5}$ & $4.50 \times 10^{6}$& $8.31 \times 10^{4}$ & $4.38 \times 10^{5}$ & $8.27 \times 10^{3}$ & $4.36 \times 10^{4}$ \\
 \hline
 50 & $1.21 \times 10^{4}$ & $6.12 \times 10^{4}$ & $1.84 \times 10^{3}$ & $2.30 \times 10^{3}$ & 175 & 886 & 17 & 88 \\
 \hline
 100 & 902 & $4.32 \times 10^{3}$ & 137 & 657 & 13 & 62 & 2 & 6 \\
 \hline
 200 & 77 & 335 & 11 & 50 & 2 & 4 & 2 & 2 \\
 \hline
 400 & 8 & 31 & 2 & 4 & 2 & 2 & 2 &  2 \\
 \hline
 600 & 2 & 8 & 2 & 2 & 2 & 2 & 2 &  2 \\
 \hline
 800 & 2 & 3 & 2 & 2 & 2 & 2 & 2 &  2 \\ 
 \hline
 1000 & 2 & 2 & 2 & 2 & 2 & 2 & 2 &  2 \\ [1ex] 
 \hline
\end{tabular}
\end{center}
\label{DaugmanLikeBound}
\end{table*}

Looking once again at the fitted chi-square imposter distributions for Relative Entropy, we can utilize our Daugman-like bound from (\ref{eq:Max_Pop_Bound_Daugman}) to find the maximum population the datasets can achieve at a given image quality of data and a fixed recognition error, $\delta$. To begin, the cumulative of our fitted distributions, from Section \ref{sec:FittingImposters}, $\chi^2(\tau)$ from 0 to $\tau$ are found, where $\tau$ is a given relative entropy metric interpreted as a distance between two codewords. Now using these cumulatives, we can find the FMR from equation \ref{eq:Max_Pop_Bound_Daugman} given a particular relative entropy value, $\tau,$ and a fixed recognition error, $\delta$ (which is varied similar to Daugman~\cite{Daugman2003}, to reflect recognition errors of 50\%, 10\%, 1\%, and .1\%). Figure \ref{fig:CASIA_BATH_Daugman} shows the resulting bounds for the CASIA-IrisV3 Interval and BATH datasets, and is illustrated for fixed values of the relative entropy in Table~\ref{DaugmanLikeBound}. Note that $\tau$ plays the role of the minimum possible distance allowed between two codewords that belong to two different classes. This implied that the distance between two codewords from the same class is approximately $\tau/2$ or less, which is achievable only if enrolled and query data in the form of codewords are of high quality.

Looking at Table~\ref{DaugmanLikeBound}, we can see that as the relative entropy (the distance between codewords), $\tau,$ increases, the maximum population obtainable, given a certain recognition probability, decreases exponentially. Looking ideally at a recognition error of $\delta = 0.001$ and  $\tau = 1$, the maximum population obtainable by the CASIA-IrisV3 Interval dataset is $2.43 \times 10^7$ and for the BATH dataset, it is $1.30 \times 10^8$. This result follows intuition because the BATH dataset is of greater quality than the CASIA-IrisV3 Interval dataset, therefore it can sustain  20\% more classes. To turn an idealized possibility into reality, a low value of relative entropy (minimum distance between two codewords of two different classes or maximum distance between codewords of the same class) can be achieved by ensuring that the quality of all data is as high as possible, which can be accommodated using modern cameras capable of collecting hundreds of frames over a short interval of time followed by the application of a bulk of signal/image processing techniques.

%%%%%%%%%%%%%%%%%%%
\section{Conclusion}
\label{sec:summary}
%%%%%%%%%%%%%%%%%%%
%
This work presents a new methodology to find the maximum population of an iris database in both a closed system perspective, Sphere Packing Bound, and an enrollment perspective, the Daugman-like Bound. Within the presented framework, a measure of the maximum population is also dependent on the quality of the iris images contained in the datasets themselves. For the Sphere Packing Bound, as the noise variance, $N,$ in the iris dataset increases, the maximum population decreases exponentially. The Daugman-like Bound presents a similar measure of iris quality on the constraint of the distance measure of relative entropy between two classes' power spectral densities, which is dependent on the noise and distortions present in the images. The size of the enrolled population can be increased by choosing a smaller value of the relative entropy (the distance between any two classes), which is achievable when the quality of data is improved. This can be attained due to the more modern data acquisition techniques and data processing applied to the dataset itself. 

In the future, we would like to develop an encoding technique bypassing the vectorization step and directly mapping iris images of different iris classes to Gaussian codewords with i.i.d. entries, each zero mean and variance $P,$ then see how this mapping affects our maximum attainable population. In conclusion, with the application of the methodology presented above, researchers can better understand the dependence of the capacity of their datasets on the data quality. An appealing approach to achieving this goal is offered by Nguyen et al.~\cite{Nguyen2020} whose work presents a constrained design of Deep Iris Networks. This will be the approach that we intend to pursue in our empirical valuation of iris biometrics capacity. 
%estimating positions and the noise radii of iris classes in a given feature space with the ultimate goal of applying a sphere packing bound.  %feature noise of a pair of iris classes estimating the distance between two distinct iris for us to explore. 

%Even though our approach to evaluating the capacity of the iris biometric is based on a Gabor filter-based encoding method. The concept of capacity also can be applied to other encoding methods or classifiers. An appealing approach in recent years is offered by Nguyen et al.~\cite{Nguyen2020} whose work presents a constrained design of Deep Iris Networks. We intend to pursue those approaches next to show how iris biometrics capacity can be different when a Neural Network classifier is used.

%One potential approach to estimate the capacity of Iris biometrics is 
%From the theory and implementation presented, the maximum population of an iris database can be found, dependent on image quality, by both a closed system perspective, Sphere-Packing Bound, and an enrollment perspective, the Daugman-like Bound. 

%\color{purple} Add comments about future work here... \color{black}

\section*{Acknowledgement}
This material is based upon work supported by the Center for Identification Technology Research and the National Science Foundation under Grant No. 1650474 and 1650503.

%%%%% References %%%%%

\bibliography{main}   % bibliography data in main.bib
\bibliographystyle{plain}   % makes bibtex use spiejour.bst

%%%%% Biographies of authors %%%%%

%\vspace{2ex}\noindent\textbf{First Author} is an assistant professor at the University of Optical Engineering. He received his BS and MS degrees in physics from the University of Optics in 1985 and 1987, respectively, and his Ph.D. degree in optics from the Institute of Technology in 1991.  He is the author of over 50 journal papers and has written three book chapters. His current research interests include optical interconnects, holography, and optoelectronic systems. He is a member of SPIE.

%\vspace{1ex}
%\noindent Biographies and photographs of the other authors are not available.

%\listoffigures
%\listoftables

\begin{IEEEbiography}[{\includegraphics[width=1in,height=1.25in,clip,keepaspectratio]{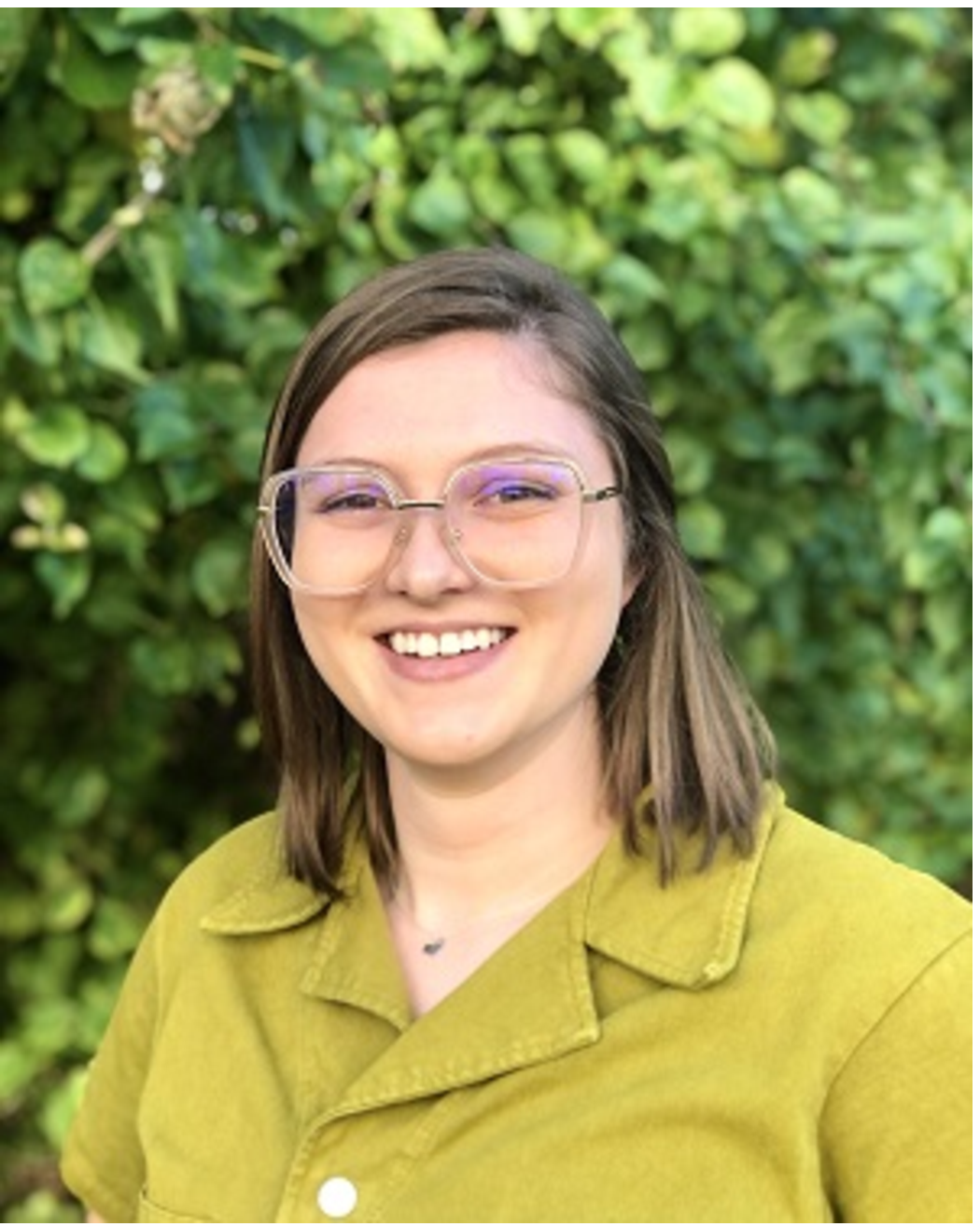}}]{Katelyn M. Hampel} (M'19) Katelyn Hampel received a B.S. degree in Electrical Engineering from West Virginia University, Morgantown, WV. She is currently working toward an M.S. degree in Electrical Engineering from West Virginia University. Her research interests include signal and image processing and detection and estimation theory in biometrics and power systems. Her recent research has focused on the uniqueness of the iris as a biometric.
\end{IEEEbiography}

\begin{IEEEbiography}
[{\includegraphics[width=1in,height=1.25in,clip,keepaspectratio]{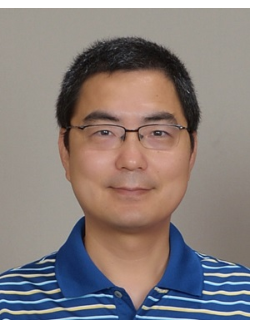}}]{Jinyu Zuo} (M'05--SM'20) 
Jinyu Zuo earned his M.S. degree in Computer Science and Technology at Tsinghua University, Beijing, China in 2004. He then pursued his
Ph.D. in Electrical Engineering at West Virginia University (WVU). After 8 months as a Post-Doctoral Fellow at WVU, Jinyu joined Polar Rain Inc.
as a Senior Research Scientist. Since 2014, he has been working at Symantec (Broadcom) as a Research Engineer. His research interests include image processing, pattern recognition, computer vision, and biometrics. His recent research has focused on detecting sensitive information for Data Loss Prevention (DLP). He has been an IEEE member since 2005 and his key research contributions are documented in 4 US patents, 5 peer-reviewed journal publications, and 16 conference presentations which have accumulated more than 1000 citations in total.
\end{IEEEbiography}

\begin{IEEEbiography}
[{\includegraphics[width=1in,height=1.25in,clip,keepaspectratio]{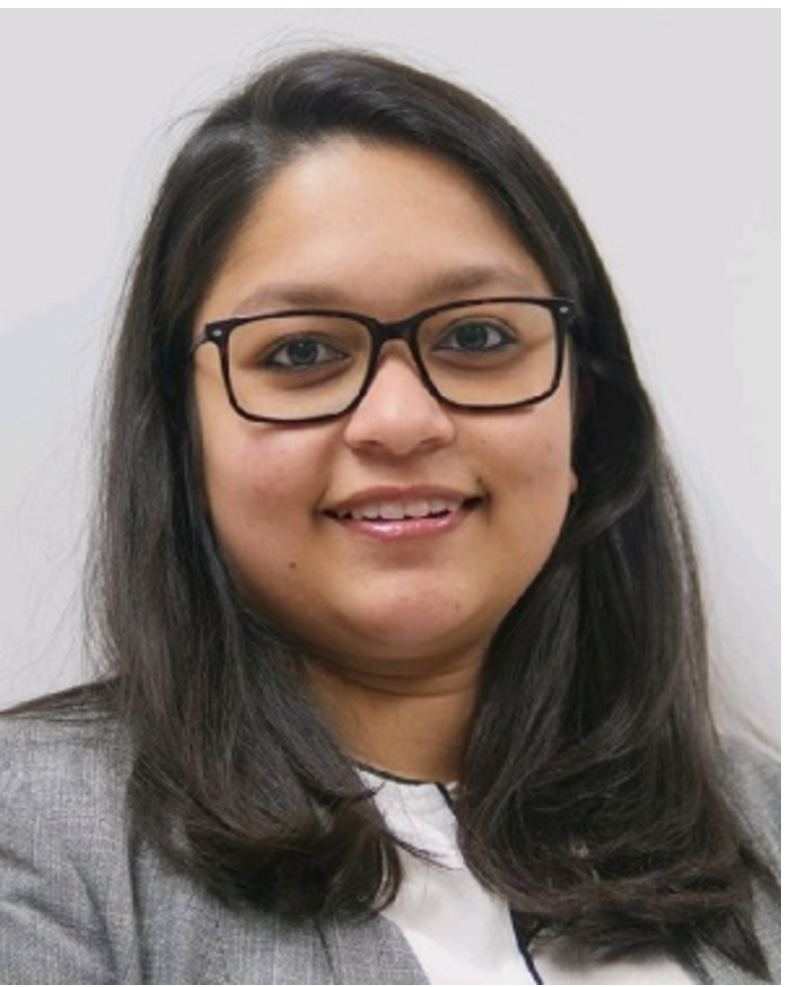}}]{Priyanka Das} 
Dr. Priyanka Das is a Data Scientist with HID Global. Her research area of interest includes biometrics, computer vision, data
science, and human-computer interaction. She
received her doctoral degree in the Department
of Electrical and Computer Engineering at Clarkson University, USA with a focus on iris biometrics. She completed her ME in Bioscience and
Engineering from Jadavpur University, India in
2016 and her Bachelor’s in Biomedical Engineering in 2014
\end{IEEEbiography}

\begin{IEEEbiography}[{\includegraphics[width=1in,height=1.25in,clip,keepaspectratio]{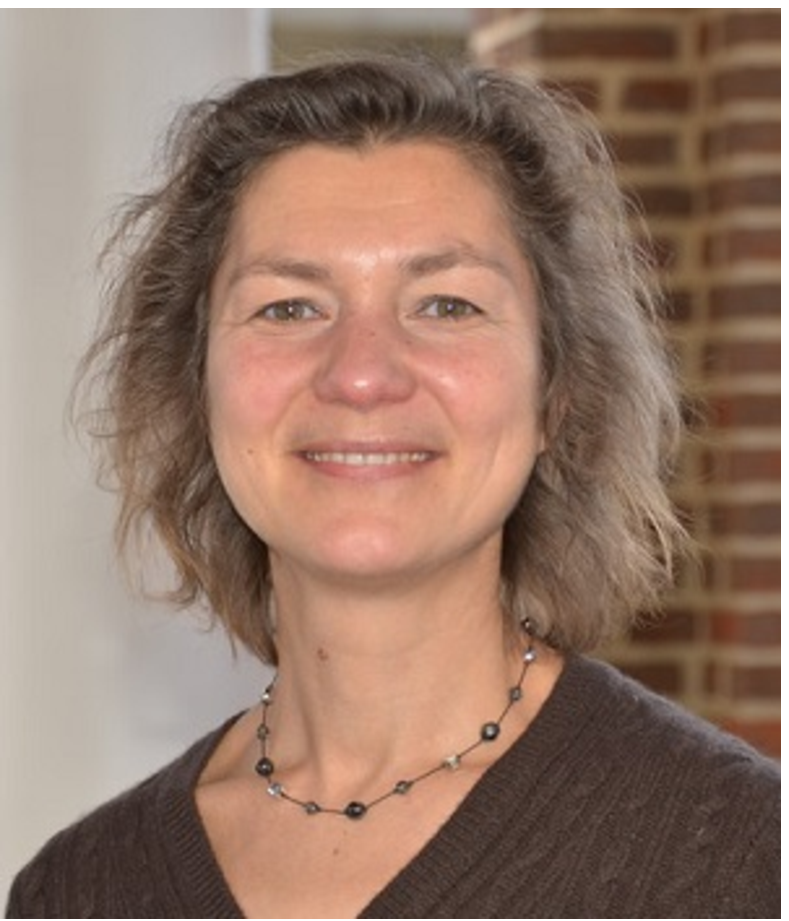}}]{Natalia A. Schmid}  
 Natalia A. Schmid received her Ph.D. in engineering from the Russian Academy of Sciences and her DSc in electrical engineering from Washington University in St. Louis, Missouri. She is currently a Professor in the Department of Computer Science and Electrical Engineering at West Virginia University. Her research and teaching interests include detection and estimation, digital and statistical signal processing, information theory, and machine learning. Her latest research is centered around radio astronomy, biometrics, and digital forensics being funded by the National Science Foundation and the National Aeronautics and Space Administration. She is the author of over 100 peer-reviewed journal and conference papers.  She is a member of IEEE. 
\end{IEEEbiography} 

\begin{IEEEbiography}
[{\includegraphics[width=1in,height=1.25in,clip,keepaspectratio]{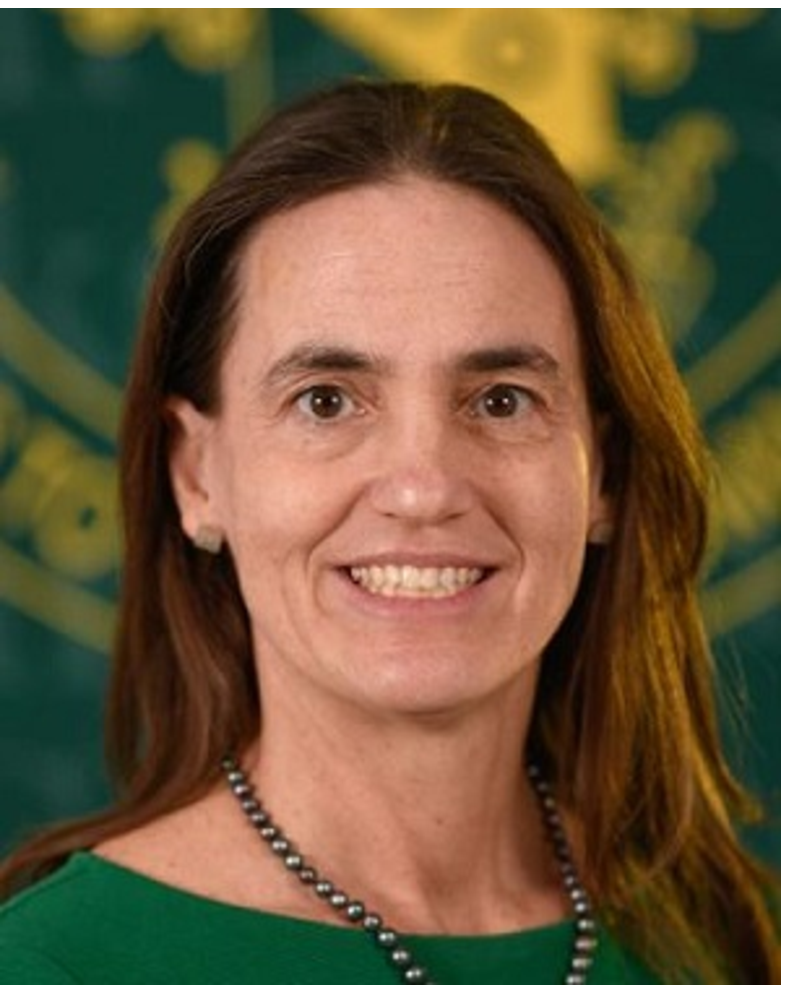}}]{Stephanie Schuckers} 
Dr. Stephanie Schuckers is the Paynter-Krigman Endowed Professor
in Engineering Science in the Department of
Electrical and Computer Engineering at Clarkson University and serves as the Director of
the Center for Identification Technology Research (CITeR), a National Science Foundation
Industry/University Cooperative Research Center. She received her doctoral degree in Electrical Engineering from The University of Michigan. Professor Schuckers' research focuses on
processing and interpreting signals which arise from the human body.
Her work is funded by various sources, including the National Science
Foundation, the Department of Homeland Security, and private industry,
among others. She has started her own business, testified for the US
Congress, and has over 100 other academic publications.
\end{IEEEbiography}

\begin{IEEEbiography}
[{\includegraphics[width=1in,height=1.25in,clip,keepaspectratio]{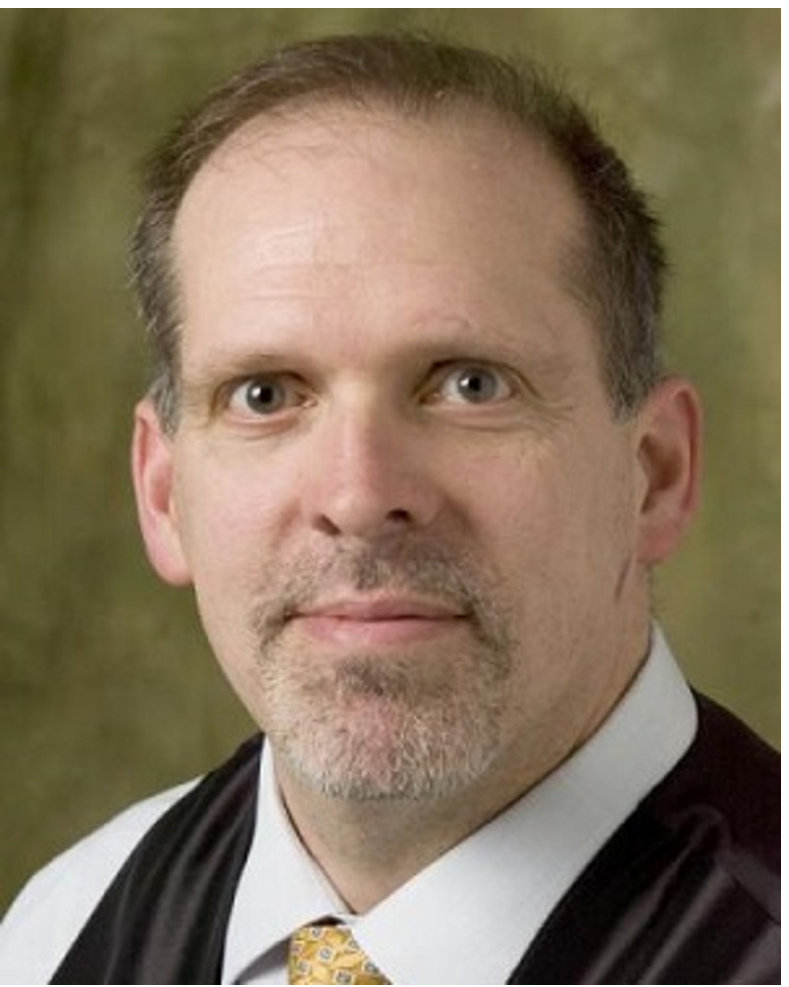}}]{Joseph Skufca} 
Joseph Skufca is the Chair of
Mathematics at Clarkson University. He received
a bachelor of science degree from the United
States Naval Academy in 1985. He served 20
years in the Submarine Force, with sea tours
aboard both fast attack and ballistic missile submarines of both the Atlantic and Pacific Fleets.
He retired from active duty in 2005. He earned
his Master of Science (2003) and Ph.D. (2005)
degrees in applied mathematics from the University of Maryland, College Park. His research
stretches across a broad spectrum of applied mathematics, with a
particular focus on applied mathematical modeling, both with analytic
and data methods.
\end{IEEEbiography}

\begin{IEEEbiography}
[{\includegraphics[width=1in,height=1.25in,clip,keepaspectratio]{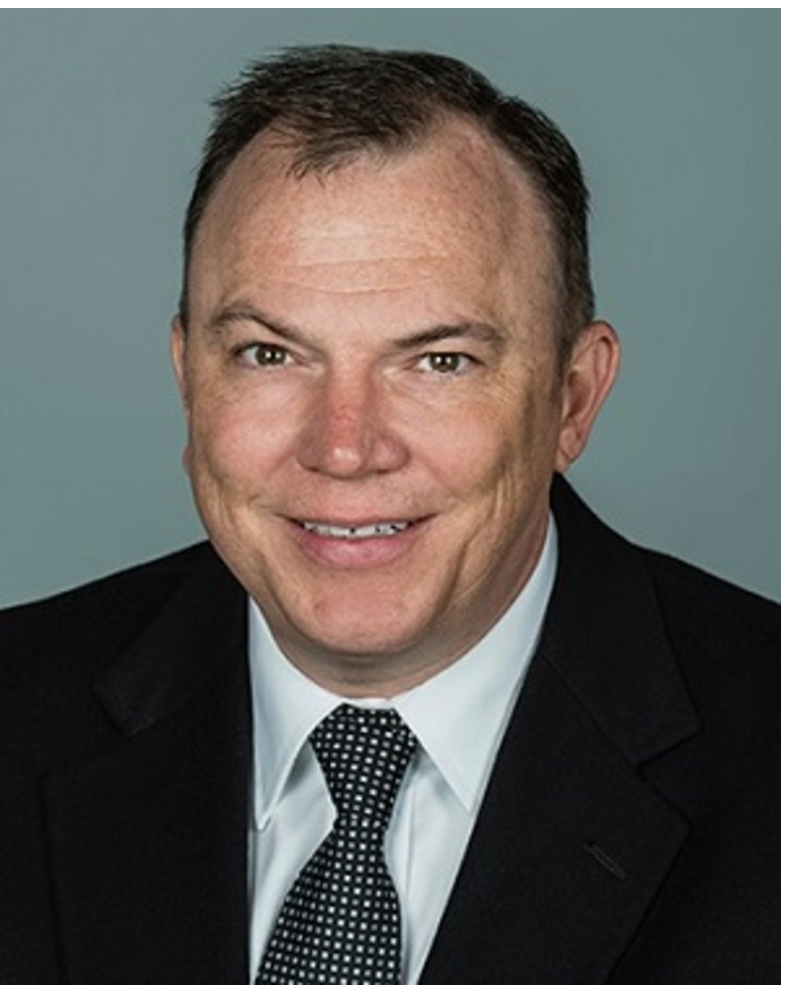}}]{Matthew C. Valenti} 
Matthew Valenti is a Professor
in the Lane Department of Computer Science
and Electrical Engineering at West Virginia University. Dr. Valenti’s research and teaching interests are in the areas of communication theory,
wireless networks, error control coding, cloud
computing, and secure multimodal biometrics.
He received B.S. and Ph.D. degrees from Virginia Tech and an M.S. from the Johns Hopkins
University. He previously worked as an Electronics Engineer at the U.S. Naval Research Laboratory. Dr. Valenti serves as Director of the West Virginia University site in
the Center for Identification Technology Research (CITeR), an NSF industry/university cooperative research center (I/UCRC). He is a recipient
of the 2019 IEEE MILCOM Award for Sustained Technical Achievement
and the 2021 IEEE Communications Society Communication Theory
Committee Outstanding Service Award. Dr. Valenti is registered as a
Professional Engineer (P.E.) in the state of West Virginia and is a Fellow
of the IEEE.
\end{IEEEbiography}

%\EOD

\end{spacing}

\end{document}